\documentclass[journal]{IEEEtran}
\usepackage[normalem]{ulem}
\usepackage[pagebackref=true,breaklinks=true,colorlinks,bookmarks=false]{hyperref}
\usepackage[pdftex]{graphicx}
\usepackage{array}
\usepackage{xcolor}
\usepackage{multirow}
\usepackage{cite}
\usepackage{amsmath}
\usepackage[caption=false,font=footnotesize]{subfig}
\usepackage{url}
\usepackage{amsfonts}
\newcommand{\RNum}[1]{\uppercase\expandafter{\romannumeral #1\relax}}
\usepackage{float}
\usepackage{pifont}

\usepackage{adjustbox}

\newlength\fsdurthree

\begin{document}

\title{Unsupervised network for low-light enhancement }

\author{Praveen Kandula, Maitreya Suin, and~A. N. Rajagopalan, \textit{Senior Member, IEEE}}

\markboth{Journal of \LaTeX\ Class Files,~Vol.~14, No.~8, August~2015}%
{Shell \MakeLowercase{\textit{et al.}}: Bare Demo of IEEEtran.cls for IEEE Journals}

\makeatletter
\def\ps@IEEEtitlepagestyle{
  \def\@oddfoot{\mycopyrightnotice}
  \def\@evenfoot{}
}

\def\mycopyrightnotice{
  {\footnotesize
  \begin{minipage}{\textwidth}
  \centering
  Copyright~\copyright~2023 IEEE. Personal use of this material is permitted. However, permission to use this  \\ 
  material for any other purposes must be obtained from the IEEE by sending a request to pubs-permissions@ieee.org.
  \end{minipage}
  }
}

\maketitle
\begin{abstract}
Supervised networks address the task of low-light enhancement using paired images. However, collecting a wide variety of low-light/clean paired images is tedious as the scene needs to remain static during imaging. In this paper, we propose an unsupervised low-light enhancement network using context-guided illumination-adaptive norm (CIN). Inspired by coarse to fine methods, we propose to address this task in two stages. In stage-\RNum{1}, a pixel amplifier module (PAM) is used to generate a coarse estimate with an overall improvement in visibility and aesthetic quality. Stage-\RNum{2} further enhances the saturated dark pixels and scene properties of the image using CIN. Different ablation studies show the importance of PAM and CIN in improving the visible quality of the image. Next, we propose a region-adaptive single input multiple output (SIMO) model that can generate multiple enhanced images from a single low-light image. The objective of SIMO is to let users choose the image of their liking from a pool of enhanced images. Human subjective analysis of SIMO results shows that the distribution of preferred images varies, endorsing the importance of SIMO-type models. Lastly, we propose a low-light road scene (LLRS) dataset having an unpaired collection of low-light and clean scenes. Unlike existing datasets, the clean and low-light scenes in LLRS are real and captured using fixed camera settings. Exhaustive comparisons on publicly available datasets, and the proposed dataset reveal that the results of our model outperform prior art quantitatively and qualitatively.
\end{abstract}
\IEEEpeerreviewmaketitle

\section{Introduction and related works}
\label{Sec:Introduction}
 
\IEEEPARstart{L}{ow}-light images suffer from poor visibility due to a small number of photons hitting the image sensor, high noise due to camera ISO, and poor contrast. Given a low-light image, the objective of low-light enhancement is to improve the overall aesthetic quality of the image. Along with human visibility preference, low-light enhancement is indispensable for numerous computer vision tasks like object recognition \cite{rajagopalan2005background}, segmentation, remote surveillance, and many others. Several works \cite{trad1, trad2, trad3, DICM, trad6, SID, enlightengan, LLEtrad1, LLEtrad2} have been proposed in literature to address the task of low-light image enhancement. A brief discussion of the same is given below.

\textbf{Learning-based methods:} Deep learning methods have given remarkable results for different image restoration tasks like deblurring \cite{mohan2018divide, vasu2018non, mohan2021deep, nimisha2018unsupervised, vasu2018nonblind, purohit2018learning, mohan2019unconstrained, rao2014harnessing, rajagopalan1998optimal, paramanand2011depth, purohit2020region, purohit2019bringing, vasu2017local, nimisha2018generating}, dehazing \cite{dehaze2, ancuti2019ntire, purohit2019multilevel, mandal2019local}, inpainting \cite{inpaint2, suin2021distillation, bhavsar2012range}, enhancement \cite{kandula2023illumination}, super-resolution \cite{vasu2018joint, suin2020degradation, rajagopalan2003motion, nimisha2021blind, purohit2021spatially, suresh2007robust, bhavsar2010resolution, purohit2020mixed}, bokeh rendering \cite{purohit2019depth} etc., and many others. Likewise, several supervised networks are proposed for low-light image enhancement \cite{ mohit, SID, DRBN, retinex, LLEsuper1, LLEsuper2, LLEsup3, el2020aim} using pixel-wise loss functions. Among them, \cite{SID} proposed SID dataset and end-to-end learning mechanism for low-light enhancement. Recently, \cite{mohit} used an efficient architecture by performing most of the computational operations at higher scales. Although these algorithms generate state-of-the-art results, a significant drawback of these methods is the need for paired training data. For instance, the dataset of \cite{SID} is collected by changing the ISO of the camera while fixing the object and camera position to ascertain that low-light and corresponding clean images are exactly aligned. Also, a low-light image may not have a unique well-lit ground truth, as the preferred lighting condition often depends on the user's choice. To address the need for large amounts of training data, \cite{iitk}, and \cite{zero-shot} proposed zero-shot techniques for low-light enhancement. However, a major shortcoming of these methods is that they are extremely slow and are not suitable for practical applications. To relax the need for paired data, several unsupervised models \cite{CycleGAN, contrastiveCycle, UEGAN, enlightengan, Zero-DCE} have been proposed. CycleGAN \cite{CycleGAN} is a seminal work to transfer images from one domain to another using adversarial and cycle-consistency losses.\cite{UEGAN} improves the aesthetic quality of images using fidelity and adversarial losses. Recently, \cite{general_purpose_unsupervised} proposed an unsupervised restoration network that can handle multiple distortions like rain, blur, and low-light, in input images. Very recently, \cite{enlightengan} proposed Enlighten-GAN (EnGAN) to handle dark images using VGG \cite{vgg} and adversarial losses similar to \cite{UEGAN}. 

Spatially varying illumination levels are common during night-time photography where the object of interest is well-lit, and other regions are dark. Despite recent advancements, prior art  \cite{LLEsuper1, LLEsuper2, LLEsup3, zero-shot, general_purpose_unsupervised} fails to handle these spatially-varying scenarios in low-light. To effectively tackle this challenging problem, we propose context-guided illumination adaptive norm that identifies varying illumination levels in low-light images and then use this crucial information for enhancement.

\begin{figure*}[t]
\setlength{\tabcolsep}{1pt}
\scriptsize
\centering
\begin{tabular}{cc}
\includegraphics[width=\linewidth]{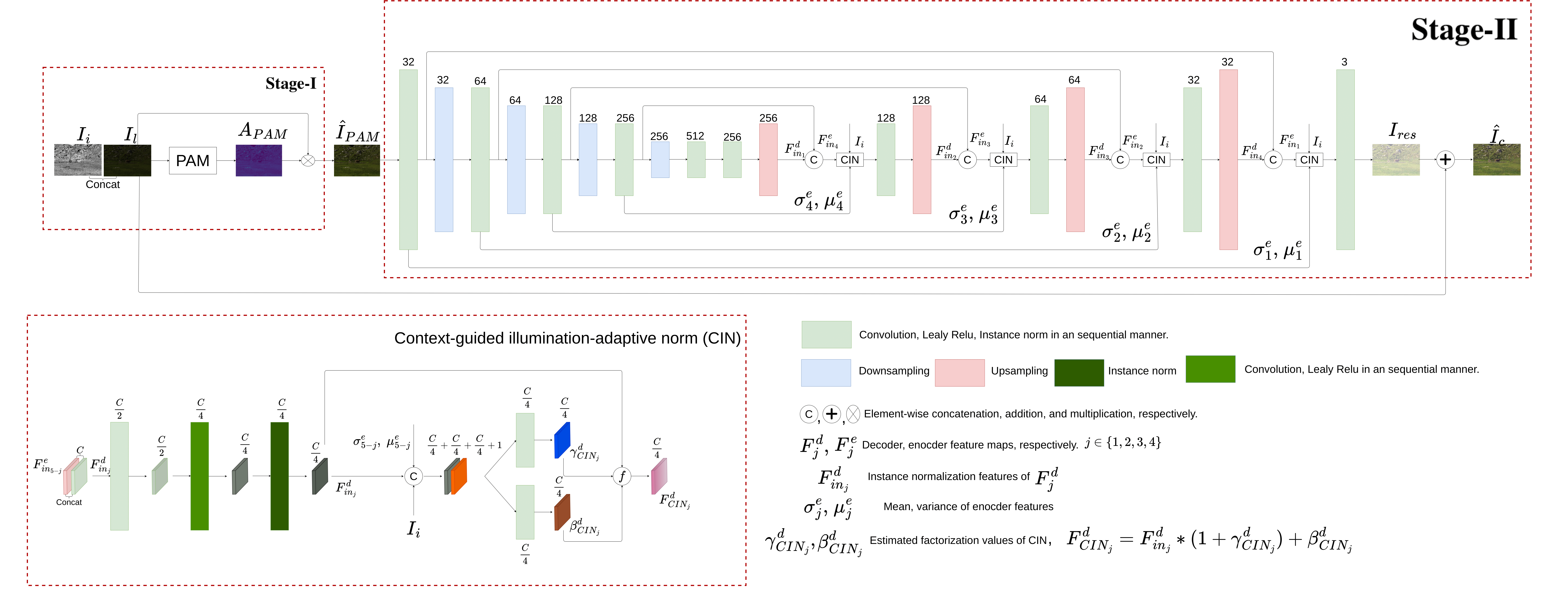} 
\end{tabular} 
\caption{Proposed network architecture for unsupervised low-light enhancement. The proposed model enhances the low-light images in two stages. Given a low-light image $I_l$ and illumination map $I_i$, stages-\RNum{1} uses PAM to generate an amplification factor $A_{PAM}$.    $I_i$ is multiplied with $A_{PAM}$ to generate $\hat{I}_{PAM}$, a coarse initial estimate with an overall improvement in visibility compared to $I_l$. Stage-\RNum{2} further enhances the dark regions of stage-\RNum{1} results using anencoder-decoder architecture and CIN. The generated residual ($I_{res}$) from stage-\RNum{2} is added to $I_l$ to get the final enhanced image $\hat{I}_c$.}

\label{fig: proposed_method}
\end{figure*}

In this paper, we propose an unsupervised low-light enhancement network that can adaptively process each pixel depending on its content, the lighting condition, and the global scene context. The adaptiveness originates from a dynamically generated per-pixel amplification map and spatially-varying feature normalization operation, which gradually shift the low-light image's feature statistics to that of a brightly-lit image. To handle the ill-posedness of the task, we design a coarse-to-fine framework, where in stage-\RNum{1}, we coarsely improve the illumination level using a pixel-level amplifier module (PAM). Given a low-light image, PAM generates an amplification factor for each pixel, which is multiplied with the input image. Although the contents of the stage-\RNum{1} results are visible to the naked eye, they often lack finer spatial details or global consistency of colour and illumination levels. This is addressed in the UNet-based Stage-\RNum{2}, which takes this coarse output along with the low-lit image as input, and refines it to produce the final enhanced image. We extract high-level global scene information from the encoder layers of Stage-\RNum{2}. In the decoder, we introduce context-guided illumination adaptive norm (CIN), which utilizes the contextual (global scene) information from the encoder, and transforms the decoder features maps in a pixel-wise manner, aiming to reach the data distribution of well-lit images. Our ablation studies demonstrate the need for two stages and the utility of CIN for the enhancement task.

Our main contributions are summarized below:
\begin{itemize}
 \item We design a two-stage unsupervised framework for enhancing low-light images. The first stage improves the overall visibility using a pixel-wise amplification module, and the second stage further enhances the image using context-adaptive illumination-guided norm (CIN). Different ablation studies show the importance of amplification module and CIN in the enhancement process.
 \item Comparison experiments reveal that the results of our model outperform prior unsupervised methods both qualitatively and quantitatively.
\end{itemize}

\section{Proposed Method}
\noindent\textbf{Stage-\RNum{1}:} We investigate the use of the amplifier model for LLE in unsupervised settings. Existing supervised networks \cite{SID, amp1, mohit2} make use of ground truth (reference) exposure settings for training the amplifier model. Training unsupervised networks with amplifier model is challenging due to the unavailability of paired images resulting in inability to use exposure settings. Instead of using a global amplifier module, we employ a pixel-wise amplifier model (PAM) as it is better-suited for unsupervised LLE. Pixel-wise amplifier module (PAM) is designed using a series of convolution blocks and non-linear activations. Specifically, PAM contains four convolution layers with ReLU after every convolution except for the last layer, which uses sigmoid activation. As shown in Fig. \ref{fig: proposed_method}, PAM uses  both $I_l \in \mathbb{R}^{3 \times H \times W}$   and $I_i \in \mathbb{R}^{H \times W}$ to generate initial amplification estimate $A_i  \in \mathbb{R}^{H \times W}$ where $H, W$ are the height and width of the image. $I_i$ is multiplied with $A_i$ to generate the enhanced image. The same can be written as 
\begin{equation}
\begin{split}
    A_i = PAM({I_i, I_l}), \\
    I^{PAM}_l = I_l \otimes A_i
\end{split}
\end{equation}
where $I^{PAM}_l$ is the enhanced image after stage-\RNum{1}, and $\otimes$ represents pixel-wise multiplication.



\noindent\textbf{Stage-\RNum{2}:} 
As shown in Fig. \ref{fig: proposed_method}, stage-\RNum{2} acts on the results of stage-\RNum{1}. Although the results of PAM significantly remove low-light and generate clean images, the results still suffer from saturated pixels and non-uniform contrast. We believe that the finer details are compromised as PAM acts on each pixel independently without access to context information. The objective of stage-\RNum{2} is to generate a well-lit image using context-guided illumination-adaptive norm (CIN).  which uses context information from input image features to restore the finer details while giving more attention to the darker regions using the illumination map. Stage-\RNum{2} is designed using UNet architecture with encoder-decoder as the backbone and skip connections between them. Majority of the existing works use batch/instance normalization throughout the network. The global feature statistics are used in such layers to scale the feature maps. We claim that applying a global normalization step to all the spatial locations is counter-productive, as it applies the same transformation on all the pixels irrespective of their illumination level. Instead, we employ standard instance-norms in the encoder and store each level's global feature statistics (mean and variance). While generating the enhanced image in the decoder, we introduce CIN for illumination-adaptive transformation of each pixel. Our intuition is to extract the global contextual information in the encoder and then transform different regions adaptively in the decoder utilizing the global scene information. The encoder module uses four blocks, with each block comprising a convolutional layer followed by  ReLU, instance normalization, and max pool sequentially. Similarly, the decoder is also designed using four blocks, with each block comprising convolutional layer+upsampling,  ReLU, followed by CIN, sequentially. A detailed discussion of CIN is given below.

Given a low-light image ($I_l$) and stage-\RNum{1} output, the stage-\RNum{2} generates a residual image ($I_{res}$) which is then added back to $I_l$ for the final clean or well-lit image ($\hat{I}_c$) as shown in Fig. \ref{fig: proposed_method}. The same can be written as 
\begin{equation}
\label{Eq: final}
    \hat{I}_c = I_{res} + I_l
\end{equation}




\noindent\textbf{Loss functions:} We use adversarial \cite{GAN} and VGG \cite{vgg} loss to train stage-\RNum{1} and stage-\RNum{2} of our proposed model. The objective of adversarial loss is to generate a clean image from a low-light input. Given a low-light image $I_l$, the adversarial loss to transfer image from low-light to clean domain is as follows.
\begin{equation}
\label{Eq: adverserail_loss}
\begin{split}
 {L}_{adv} =  
 \mathbb{E}[\log (D_{adv}(I_r)] +
  \mathbb{E} [\log(1 - D_{adv}(\hat{I}_c)]  
\end{split}
\end{equation}
where $I_r$ is the image sampled from clean set, $\hat{I}_c$ is the clean image generated from $I_l$,  and $D_{adv}$ is the discriminator network to differentiate between clean and fake samples.

Although $L_{adv}$ transfers the low-light image to a clean domain, it does not necessarily ensure that the contents between $I_l$ and $\hat{I}_{c}$ remain the same. We employ the widely used perceptual loss to resolve this, which ensures the VGG features for both low-light and corresponding restored images remain the same. The perceptual loss between $I_l$ and $\hat{I}_{c}$ can be written as 
\begin{equation}
\label{Eq: percep_loss}
    L_{perc} = ||F_{vgg}(I_l) - F_{vgg}(\hat{I}_{c})||
\end{equation}
where $F_{vgg}$ refers to feature extraction from a pretrained VGG model, and $I_l$, $\hat{I}_c$ are the low-light and the estimated clean image, respectively. 

The total function  to train our model is the combination of Eq. \ref{Eq: adverserail_loss} and Eq. \ref{Eq: percep_loss}, and is written as 
\begin{equation}\
\label{Eq: total_loss}
    L_{total} = \omega_1{L}_{adv} + \omega_2 L_{perc}
\end{equation}
where $\omega_1$ and $\omega_2$ are the trade-off weights to balance adversarial and perceptual losses.

In our experiments, we used VGG features from $ReLU5\_3$ for $L_{perc}$. Unlike existing low-light methods \cite{SID, mohit2}, we do not employ individual loss functions on stage-\RNum{1} output due to unavailability of reference exposure settings. Instead, the weights of stage-\RNum{1} are optimized using Eq. \ref{Eq: total_loss} alone.

\section{Experiments}
\subsection{Implementation details}
We used NVIDIA-2080 Ti GPU to train and test our model. Both stage-\RNum{1} and stage-\RNum{2} are trained in an end-to-end manner using Eq. \ref{Eq: total_loss}. Although satisfactory results are observed around 40-50 epochs, the training is continued until 80 epochs for better convergence. We empirically found $\omega_1$ = 1 and $\omega_2$ = 0.5 for first five epochs, and then $\omega_1$ = 1 and $\omega_2$ = 1 for the remaining 75 epochs, gave best performance. Following options are used to train the network: patch size of 256 and batch size of 4, ADAM optimizer with momemtum values $\beta_1$ = 0.9 and $\beta_2$ = 0.99 for optimizing the weights. Stage-\RNum{1} is a light weight network designed using three convolutional layers with hidden sizes of 32, 32, and 1, respectively. Network architecture details of stage-\RNum{2} are shown in Fig. \ref{fig: proposed_method}.

For training, we used publicly available unpaired datasets by UEGAN \cite{UEGAN} and EnGAN \cite{enlightengan}, and the proposed low-light road scene dataset (LLRS). The authors of EnGAN \cite{enlightengan} collected an unpaired set of clean and low-light images. The dataset includes 914 low-light and 1016 well-lit images collected from different sources \cite{dataset1, dataset2, retinex}. We used the official train/test split provided by the authors. Very recently, UEGAN \cite{UEGAN} collected an unpaired set for image aesthetics improvement. The dataset includes a set of 2250 raw images treated as low-quality and another set of 2250 expert-touched images treated as target domain or high quality. The objective is to transfer the low-quality images to expert-touched ones. No paired images are present during the training phase for all the above datasets. Our network is trained separately for both EnGAN and UEGAN datsets, and the inference is done on the paired test sets provided by EnGAN \cite{enlightengan} and UEGAN \cite{UEGAN} with their respective trained models. To quantitatively evaluate our model, we used PSNR, SSIM, MSE, NIQE, BRISQUE, and PIQE. Among these, PSNR, SSIM, and MSE are reference-based metrics, i.e., a reference or ground-truth image is needed to calculate these metrics. On the other hand, NIQE, BRISQUE, and PIQE are no-reference metrics, i.e., no additional image is required to calculate these metrics. No-reference metrics are becoming popular and relevant for unsupervised methods as they can quantitatively evaluate the generated images without the need for ground-truth images.

\begin{figure*}[t]
\setlength{\tabcolsep}{1pt}
\scriptsize
\centering
\begin{tabular}{ccccccccccccc}
\includegraphics[width=0.22\linewidth]{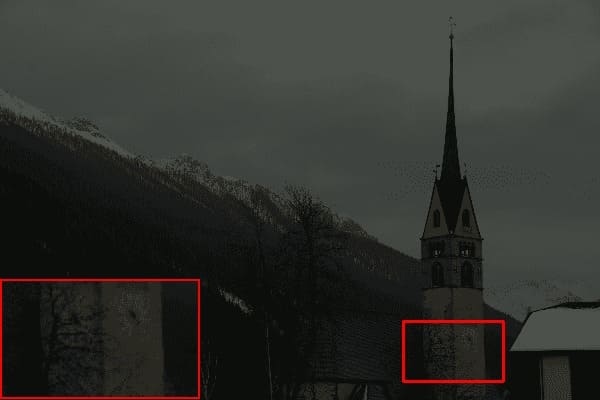} &
\includegraphics[width=0.22\linewidth]{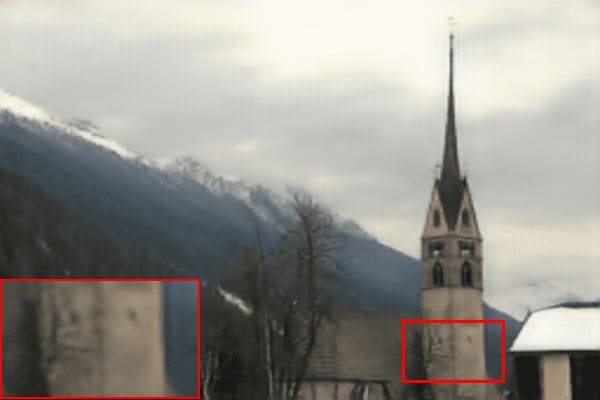} &
\includegraphics[width=0.22\linewidth]{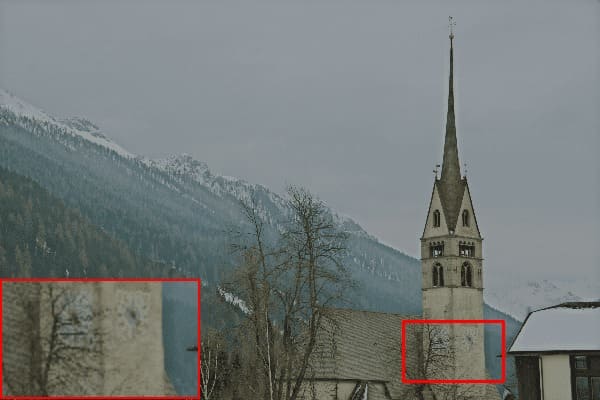} &
\includegraphics[width=0.22\linewidth]{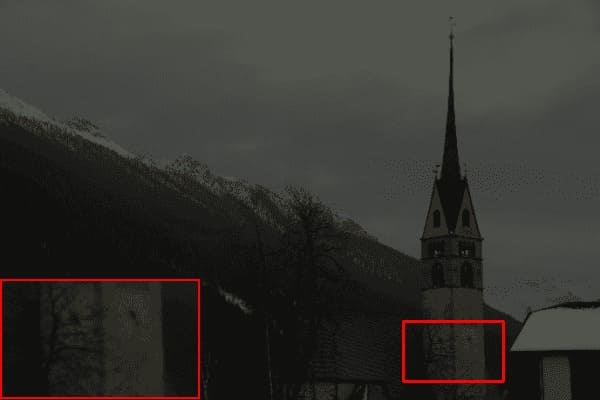} \\
Input (7.97/0.45) & Cycle GAN \cite{CycleGAN} (24.65/0.72) & Zero-DCE \cite{Zero-DCE} (15.94/0.74) & UEGAN \cite{UEGAN} (8.15/0.45) \\
\includegraphics[width=0.22\linewidth]{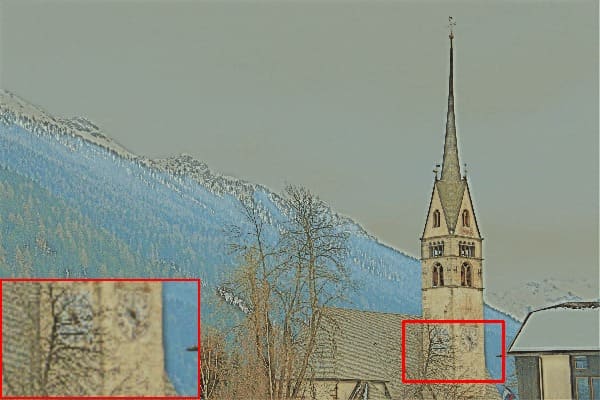} &
\includegraphics[width=0.22\linewidth]{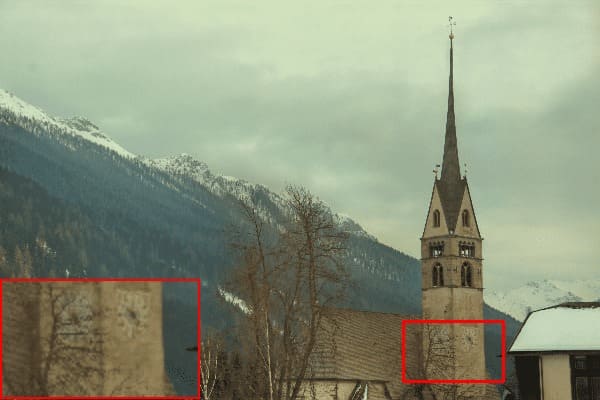} &
\includegraphics[width=0.22\linewidth]{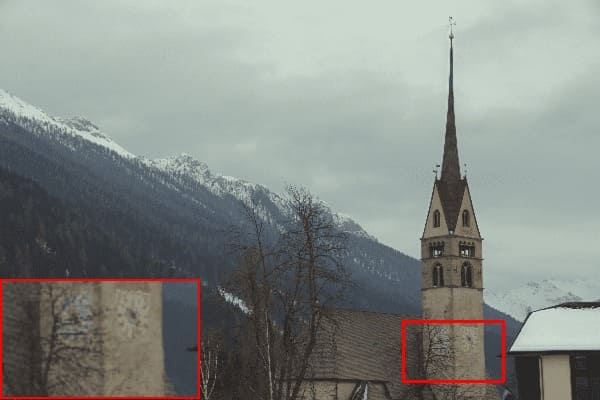} &
\includegraphics[width=0.22\linewidth]{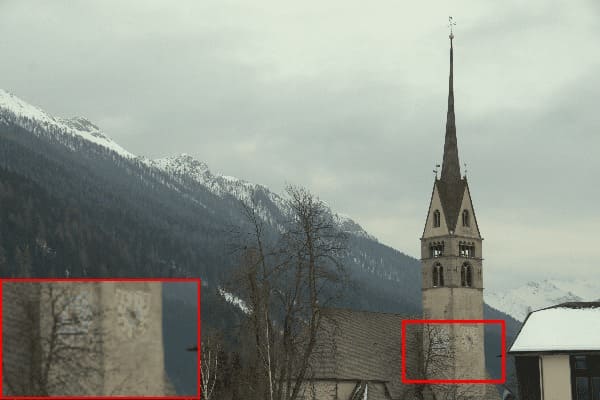} \\
Retinex-Net \cite{retinex} (13.32/0.76) & EnGAN\cite{enlightengan} (23.72/0.82) & Ours \textbf{(26.70/0.91)} & Ground truth \\
\end{tabular}\\ 

\caption{Visual comparisons with state-of-the-art unsupervised methods on low-light enhancement task. PSNR/SSIM values are provided for each result. Unnatural artifacts and colour transforms are observed in Retinex-Net \cite{retinex}, and Zero-DCE \cite{Zero-DCE}. Although low light regions are enhanced in CycleGAN \cite{CycleGAN}, the results suffer from structural incoherence reflected in SSIM values. Similarly, a yellowish tinge is present all over the result of EnGAN \cite{enlightengan}, severely affecting the visual quality and PSNR metric. On the other hand, the result of our method is free from low-light and is close to the ground truth. }
\label{fig: comparisons_EnGAN}
\end{figure*}

\subsection{Comparisons with competing methods}
We compared the results of our model with LECARM \cite{LLEtrad2}, CycleGAN \cite{CycleGAN}, DCLGAN \cite{contrastiveCycle}, UEGAN \cite{UEGAN}, EnGAN \cite{enlightengan}, ZSIR \cite{iitk}, Zero-DCE \cite{Zero-DCE}, DRBN \cite{DRBN}, Retinex-NET \cite{retinex}, RUAS \cite{RUAS}, and SCI \cite{SCI}.  Among these, LECARM is a traditional approach that generates enhanced images using a camera response model. CycleGAN and DCLGAN are unsupervised networks that transfer images from one domain to another. EnGAN and UEGAN are also unsupervised techniques for low-light enhancement and image aesthetics improvement tasks, respectively. ZSIR and Zero-DCE are zero-shot techniques for low-light enhancement that neither require paired images nor a large collection of unpaired data. RUAS \cite{RUAS} and SCI \cite{SCI} designed efficient models for unsupervised image enhancement. We also compared the results of our method with two supervised low-light enhancement networks, Retinex-NET \cite{retinex} and DRBN \cite{DRBN}. Since it is not possible to train \cite{retinex} and \cite{DRBN} using the unpaired datasets of  \cite{enlightengan} and \cite{UEGAN}, we used the pretrained models provided by the authors for testing purpose alone. CycleGAN \cite{CycleGAN}, DCLGAN \cite{contrastiveCycle}, UEGAN \cite{UEGAN}, and Zero-DCE \cite{Zero-DCE} are trained from scratch for both \cite{enlightengan} and \cite{UEGAN} datasets using the official codes provided by the authors. For EnGAN, pretrained model provided by authors is used for testing \cite{enlightengan} dataset, and the network is trained from scratch for \cite{UEGAN} dataset. \\
We performed additional quantitative evaluations of object detection results after enhancement. For this experiment, we used dark face images from DARKFACE \cite{darkface} dataset. The images are passed through the pretrained networks of the following enhancement techniques: CycleGAN \cite{CycleGAN} (trained for enhancement), EnGAN \cite{enlightengan}, RUAS \cite{RUAS}, Retinex-Net \cite{retinex}, SCI \cite{SCI}, Zero-DCE \cite{Zero-DCE}, and ours. The enhanced images from these methods are passed through a face detector \cite{face_detector}.

\section{Conclusions}
We proposed an unsupervised enhancement network to tackle spatially varying illumination levels in low-light images. Different ablation studies revealed that the proposed context-guided illumination adaptive norm (CIN) is more adaptive to varying illumination levels than batch and instance norms. Visual and quantitative comparisons demonstrate the superiority of our approach compared to prior art. Next, we proposed a region-adaptive single-input-multiple-output (SIMO), which can generate multiple enhanced images from a single low-light image. Human subjective analysis shows that preference for enhanced images is not uniform, stipulating the need for SIMO-type models for other restoration tasks. Finally, we collected a low-light road scene (LLRS) dataset that contains only real low-light and well-lit road scenes. Low light in existing datasets is either synthetically generated or due to changes in exposure settings during imaging. Differently, low-light in the LLRS dataset is due to the scene itself, as images are captured with constant camera settings. 

{\small
	\bibliographystyle{IEEEtran}
	\bibliography{egbib}
}

\end{document}